\documentclass[11pt]{article}

\usepackage[final]{acl}

\usepackage{times}
\usepackage{latexsym}
\usepackage{xspace}
\usepackage{amsthm}
\usepackage{booktabs} 
\usepackage{multirow} 
\usepackage{algorithm}
\usepackage{amsmath}
\usepackage{algorithmic}
\usepackage{changes}
\usepackage{listings} 

\usepackage{xcolor}
\usepackage{colortbl}

\usepackage{tcolorbox}
\usepackage{amssymb}
\usepackage{pifont}

\definecolor{astgreen}{RGB}{34, 139, 34}
\definecolor{astlightgreen}{RGB}{220, 255, 220}
\definecolor{textred}{RGB}{178, 34, 34}
\definecolor{textlightred}{RGB}{255, 230, 230}
\definecolor{codebg}{RGB}{248, 248, 248}
\definecolor{codeframe}{RGB}{200, 200, 200}

\tcbuselibrary{skins, breakable, listings}

\lstdefinestyle{bashstyle}{
    language=bash,
    basicstyle=\ttfamily\scriptsize,
    breaklines=true,
    backgroundcolor=\color{codebg},
    frame=single,
    rulecolor=\color{codeframe},
    xleftmargin=2pt,
    xrightmargin=2pt,
    aboveskip=4pt,
    belowskip=4pt,
}
\definecolor{poscol}{RGB}{34, 120, 34}   % green
\definecolor{negcol}{RGB}{192, 80, 77}   % red

\newcommand{\posc}[1]{\textcolor{poscol}{#1}}
\newcommand{\negc}[1]{\textcolor{negcol}{#1}}

\theoremstyle{definition}

\usepackage[T1]{fontenc}
\usepackage[utf8]{inputenc}

\usepackage{microtype}

\usepackage{inconsolata}

\usepackage{graphicx}

\usepackage{xcolor}

\newcommand{\missingcite}[1]{\textcolor{red}{[CITATION NEEDED\ifx\\#1\\\else: #1\fi]}}
\newcommand{\system}{\textsc{\textbf{CodeStruct}}\xspace}
\title{\system: Code Agents over Structured Action Spaces}

\author{
Myeongsoo Kim, Joe Hsu, Dingmin Wang \\
\textbf{Shweta Garg}, \textbf{Varun Kumar}, \textbf{Murali Krishna Ramanathan} \\
AWS AI Labs \\
\texttt{\{mysoo, hchaochu, wdimmy, shwegarg, kuvrun, mkraman\}@amazon.com}
}

\begin{document}
\maketitle
\begin{abstract}

LLM-based code agents treat repositories as unstructured text, applying edits through brittle string matching that frequently fails due to formatting drift or ambiguous patterns. We propose reframing the codebase as a \emph{structured action space} where agents operate on named AST entities rather than text spans. Our framework, \textsc{CodeStruct}, provides \texttt{readCode} for retrieving complete syntactic units and \texttt{editCode} for applying syntax-validated transformations to semantic program elements. Evaluated on SWE-Bench Verified across six LLMs, \textsc{CodeStruct} improves Pass@1 accuracy by 1.2--5.0\% while reducing token consumption by 12--38\% for most models. Models that frequently fail to produce valid patches under text-based interfaces benefit most: GPT-5-nano improves by 20.8\% as empty-patch failures drop from 46.6\% to 7.2\%. On CodeAssistBench, we observe consistent accuracy gains (+0.8--4.4\%) with cost reductions up to 33\%. Our results show that structure-aware interfaces offer a more reliable foundation for code agents.

\end{abstract}

\section{Introduction}

Large language models have enabled code agents to solve complex software engineering tasks such as repository-level bug fixing and feature implementation, as demonstrated by benchmarks like SWE-Bench~\citep{jimenez2024swebench}. However, despite their growing capabilities, current agents interact with code repositories through a fundamental abstraction mismatch: they treat programs as flat text rather than structured artifacts. Agents read files as character sequences and apply edits via specifying line numbers or string patterns. This paradigm discards the syntactic and semantic structure inherent in source code, resulting in unstructured representations that are highly error-prone.

\begin{figure}[!t]
    \centering
    \includegraphics[width=0.99\linewidth]{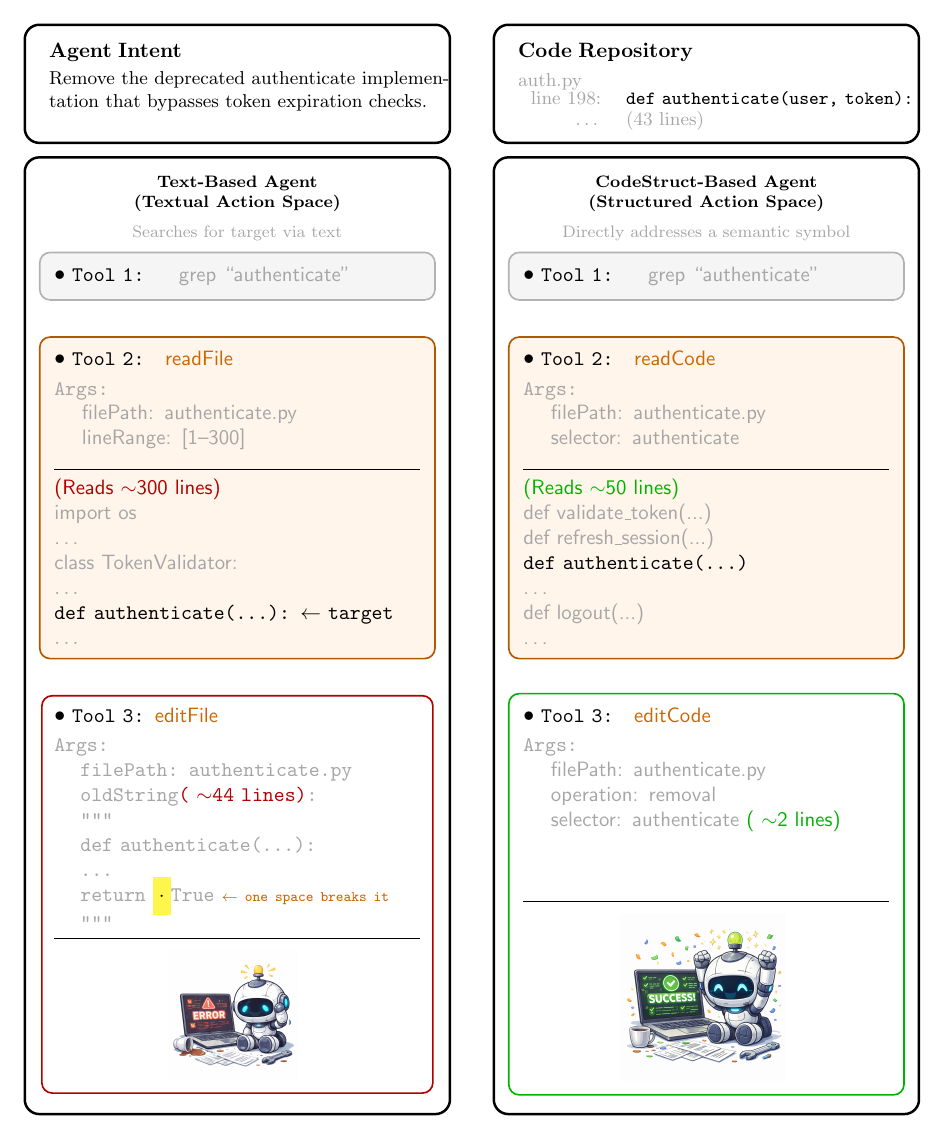}
\caption{
\textbf{Contrasting action spaces for code agents.}
Text-based agents (left) read \textasciitilde300 lines to locate a function and regenerate \textasciitilde44 lines verbatim for removal, making edits brittle to formatting changes.
\system (right) reads only the target symbol (\textasciitilde50 lines) and specifies removal in \textasciitilde2 lines via a symbol-scoped edit.
}
\vspace{-15pt}
\label{fig:overview}
\end{figure}

% This text-centric paradigm introduces critical limitations. When retrieving code, agents must choose between reading entire files, which introduces irrelevant context that degrades reasoning~\citep{shi2023distracted}, or selecting line ranges that often truncate functions mid-statement. When modifying code, string-based replacement is particularly wasteful and brittle: even minor edits require the model to regenerate significant amounts of original code verbatim. Such text-based approaches frequently encounter "no occurrence" errors when code formatting drifts, or "multiple occurrence" errors when target patterns repeat across the codebase. These systematic failures force agents into costly trial-and-error cycles, probing arbitrary line ranges until an exact match succeeds. 

% While recent systems augment text-based tools with structural summaries such as repository maps, symbol indices, or dependency graphs~\citep{yang2024sweagent,aider,zhang2024autocoderover,wang2024openhands}, these mechanisms primarily guide where agents should look rather than how they interact with code. Even for reading, agents ultimately retrieve code through line-based interfaces, selecting start and end line numbers that may truncate syntactic units or omit semantically relevant context. For writing, modifications are applied via string-based replacement or line-level patches, inheriting brittleness to formatting drift and repeated patterns. As a result, structural information is used only as auxiliary metadata, while the underlying read and write actions remain fundamentally text-based.
This text-centric paradigm introduces critical limitations for both reading and writing code. When retrieving code, agents must choose between reading entire files, which introduces irrelevant context that degrades reasoning~\citep{shi2023distracted}, or selecting line ranges that often truncate functions mid-statement. When modifying code, string-based replacement is particularly wasteful and brittle: even minor edits require the model to regenerate significant amounts of original code verbatim, and such approaches frequently encounter "no occurrence" errors when code formatting drifts, or "multiple occurrence" errors when target patterns repeat across the codebase. These systematic failures force agents into costly trial-and-error cycles.

Recent systems attempt to address these issues by augmenting text-based tools with structural summaries such as repository maps, symbol indices, or dependency graphs~\citep{yang2024sweagent,zhang2024autocoderover,wang2024openhands,aider}. However, these mechanisms primarily guide \textit{where} agents should look rather than \textit{how} they interact with code. The underlying read and write actions remain fundamentally text-based, inheriting the same brittleness and inefficiencies.

We introduce \system, a framework that grounds agent interactions in AST structure. Source code is defined by precise syntax and organized into named entities, and AST-based transformations are standard in traditional software tools~\cite{4339230, falleri2014gumtree, ragan2019comby}. However, these representations have not been adopted as the primary abstraction for LLM-based agents. Rather than operating on text spans, agents in \system reference code via \textbf{AST nodes} (e.g., \texttt{file.py::ClassName::method}) that unambiguously identify program entities regardless of line position. We provide two structure-aware primitives. The \texttt{readCode} operation retrieves complete syntactic units such as functions or classes without truncation or excess context, while \texttt{editCode} applies transformations directly to AST nodes, eliminating string-matching fragility. For node replacement, agents specify only the signature and new content, avoiding redundant regeneration of unchanged code. As illustrated in Figure~\ref{fig:overview}, operations such as deletion and duplication become atomic actions that require only a node path, yielding an efficient and structure-grounded read/write paradigm for code agents.

We evaluate \system on two complementary benchmarks, SWE-Bench Verified~\citep{jimenez2024swebench} and CodeAssistBench~\cite{kim2025codeassistbench}, across multiple language models. On SWE-Bench Verified, \system improves Pass@1 accuracy by 1.2--5.0\% for frontier models, with a 20.8 percentage point gain for a smaller model. For most configurations, these accuracy improvements coincide with reduced token consumption (12–38\%) and lower inference cost (up to 33\%). One notable exception is GPT-5-nano, which achieves a 20.8 percentage point accuracy gain at the cost of increased computation, as structured actions enable sustained exploration that would otherwise terminate in failure. On CodeAssistBench, \system consistently improves accuracy by 0.8--4.4\% across all evaluated models. These results indicate that exposing codebases as structured action spaces improves agent effectiveness and reliability across diverse code tasks.

Our contributions are:
\textbf{(1)} A structured action-space interface that bridges AST-based
program representations and LLM-based agents, exposing named semantic
entities as the primary units of code interaction;
\textbf{(2)} two structure-aware primitives (\texttt{readCode},
\texttt{editCode}) designed for LLM usability, featuring human-readable
selectors, automatic scope resolution, and syntax-validated edits that
support robust agent recovery; and
\textbf{(3)} extensive empirical evidence across two benchmarks and six language models showing that structure-aware action spaces improve both effectiveness and efficiency, with analysis revealing that gains are largest when text-interface brittleness—rather than reasoning capacity—is the dominant failure mode.\footnote{Code and data are available at \url{https://github.com/amazon-science/CodeStruct}.}

\section{Related Work}
\subsection{LLM-based Code Agents and Tools}

Recent work on LLM-based code agents focuses on enabling models to solve repository-level tasks through iterative tool use. Systems such as SWE-Agent~\cite{yang2024sweagent} and Agentless~\cite{xia2024agentless} equip agents with file-reading and text-editing tools, allowing them to explore repositories and apply patches in a multi-step manner. To improve scalability and navigation, some recent systems augment these textual tool interfaces with repository-level summaries or structure-aware retrieval mechanisms, such as file maps or symbol indices, which expose high-level information about file structure and function signatures~\cite{yang2024sweagent, zhang2024autocoderover, wang2024openhands}.

% Recent work on LLM-based code agents focuses on enabling models to solve repository-level tasks through iterative tool use. Systems such as SWE-Agent~\cite{yang2024sweagent} and Agentless~\cite{xia2024agentless} equip agents with file-reading and text-editing tools, allowing them to explore repositories and apply patches in a multi-step manner. 
% % Commercial systems and open-source agents similarly rely on textual interfaces that expose code through line-based reads and string-based edits. 
% To improve scalability and navigation, some recent systems augment these textual tool interfaces with repository-level summaries or structure-aware retrieval mechanisms, such as file maps or symbol indices, which expose high-level information about file structure and function signatures~\cite{yang2024sweagent, zhang2024autocoderover, wang2024openhands}.

While these mechanisms improve code localization and retrieval, they stop short of defining executable, structure-aware action primitives for modification, leaving reads and edits fundamentally text-based. This forces agents to reason about and manipulate structured programs indirectly, motivating the need for action abstractions that operate directly over named program entities rather than unstructured text.

\subsection{Code Search and Structural Abstractions}

A substantial body of work incorporates program structure to improve code search and understanding. Path-based models such as Code2Vec~\cite{alon2019code2vec} and PSCS~\cite{sun2020pscs} represent code using sequences of AST paths, enabling more semantically meaningful retrieval than token-based methods. Similarly, ASTNN~\cite{zhang2019astnn} and its successors decompose abstract syntax trees into statement-level subtrees for neural representation learning, with later work augmenting these embeddings using static-analysis signals for tasks such as clone detection and code classification. While these approaches demonstrate that structural information is valuable for code representation, they operate exclusively at the encoding level: AST structure is consumed as input features in single-shot prediction settings, but is not exposed as an executable action space that agents can manipulate through tree-level edits in a multi-turn workflow.

\subsection{Structure-Aware Program Repair and Code Generation}
Structural cues have also been leveraged to improve one-shot program repair and code generation. Classical systems such as DeepFix~\cite{gupta2017deepfix}, DrRepair~\cite{yasunaga2020drrepair}, and BIFI~\cite{yasunaga2021bifi} rely on ASTs or compiler feedback to reduce syntax errors and localize bugs, while grammar-based decoders explicitly enforce syntactic correctness during generation. Similarly, abstract syntax networks~\cite{rabinovich2017asn} and retrieval-augmented structural models~\cite{hashimoto2018retrieve} generate or edit code under tree- or grammar-constrained representations to respect programming-language syntax. While these approaches effectively leverage AST structure to constrain or guide prediction, they operate in a single-shot setting and do not define an executable, step-by-step action space over tree edits. As a result, models produce a complete patch or AST in one pass, rather than performing a sequence of explicit, traceable AST transformations suitable for multi-turn agent workflows.

% Structural cues have also been leveraged to improve one-shot program repair and code generation. Classical systems such as DeepFix~\cite{gupta2017deepfix}, DrRepair~\cite{yasunaga2020drrepair}, and BIFI~\cite{yasunaga2021bifi} rely on ASTs or compiler feedback to reduce syntax errors and localize bugs, and grammar-based decoders in neural program repair explicitly ensure syntactic correctness during generation. Likewise, abstract syntax networks~\cite{rabinovich2017asn} and retrieval-augmented structural models generate code in tree form to respect programming-language grammar. These methods leverage ASTs to constrain or guide single-step prediction, but do not define a step-by-step action space over tree edits; the model outputs a complete patch or AST in one pass rather than performing a sequence of explicit, traceable AST transformations.

\subsection{AST Diffing and Tree Transformations}

Work on tree diffing provides the closest analogy to our framework. 
Algorithms such as GumTree~\cite{falleri2014gumtree} compute fine-grained edit scripts between ASTs using operations such as insert, delete, update, and move, and systems such as PyGGI~\cite{an2019pyggi} adopt similar primitives for genetic improvement. 
However, these operations are primarily used for offline diffing or evolutionary search, rather than as decision-time primitives for an LLM-driven code-editing agent. Beyond diffing, several systems support structural code transformations via AST-aware rules, including Comby~\cite{ragan2019comby}, Piranha~\cite{9276556,10.1145/3656429}, and Semgrep~\cite{10.1145/3661167.3661262}. 
These tools enable precise, semantics-aware rewrites, but require transformation patterns to be specified \emph{a priori}, making them unsuitable for open-ended problem solving.

Unlike prior AST-based systems that apply transformations offline or through fixed rewrite rules, \system exposes semantic program entities as first-class, decision-time actions that an LLM agent can dynamically construct and invoke during multi-step problem solving.

% \mkr{will be good to cite comby (pldi 19), piranha (ICSE-SEIP20, PLDI 24), semgrep as other relevant works in this area of code transformations using ASTs}

% ### Source Code (auth.py)
% ```python
% # auth.py
% class AuthManager:
%     def authenticate(self, user):
%         return self._verify(user)
%     def logout(self):
%         pass
% ```

% ### AST Structure
% ```
% Module: auth.py
%     │
%     └── ClassDef: AuthManager
%             │
%             ├── FunctionDef: authenticate
%             │       └── body: [Return(...)]
%             │
%             └── FunctionDef: logout
%                     └── body: [Pass]
% ```
% Programmable \joe{what does programmable mean in this paper?}

\section{\system}
\label{sec:method}

We introduce \textsc{\system}, a structure-aware interface that exposes a codebase as a structured action space for LLM-based agents. Rather than interacting with repositories through unstructured text spans, agents using \textsc{\system} operate over named program entities derived from the abstract syntax tree (AST). This design enables agents to read and modify code via semantically grounded, executable actions with well-defined scope boundaries and structural guarantees.

\begin{figure}[t]
    \centering
    \includegraphics[width=\linewidth]{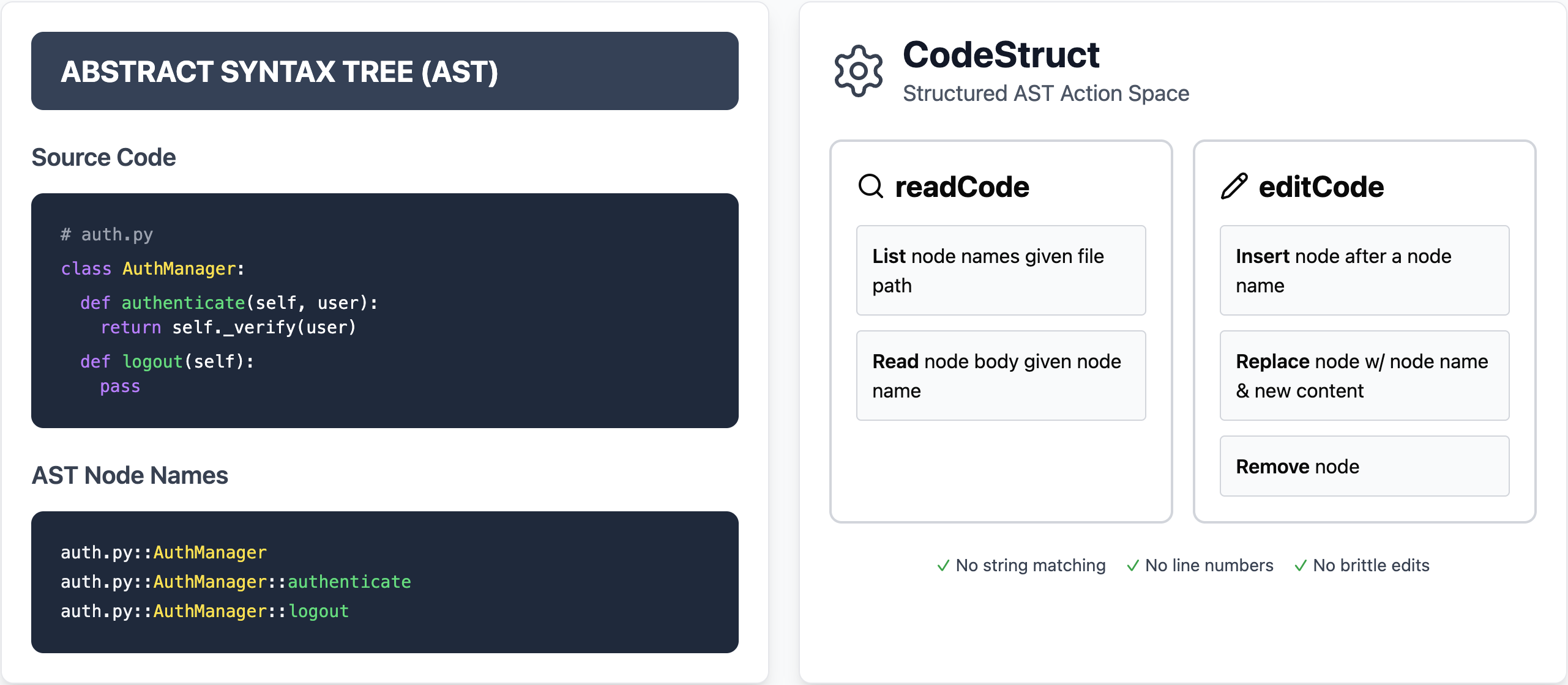}
    \caption{Overview of \textsc{\system}. Code agents interact with repositories through a structured AST action space. Source code is parsed into an AST, exposing addressable nodes. The \texttt{readCode} and \texttt{editCode} tools operate directly on these nodes, enabling structure-aware code navigation and modification without string matching, line numbers, or brittle edits.}
    \label{fig:codestruct-overview}
    \vspace{-10pt}
\end{figure}

\subsection{Structured AST Action Space}
\label{subsec:ast-action-space}

A key observation in \textsc{\system} is that source code already defines
a rich symbolic interface. Functions, classes, and methods are named program
entities with well-defined scope, boundaries, and semantics, and human developers
reason about and modify code primarily by referring to these entities by name,
rather than by line numbers or character offsets. \textsc{\system} exposes this
existing structure directly to language-model agents.

As illustrated in Figure~\ref{fig:codestruct-overview}, \textsc{\system} represents
a codebase as a structured environment defined by its abstract syntax tree (AST).
The AST serves as the environment state over which an agent operates, making named
and addressable program entities---such as files, classes, functions, and
methods---explicit and executable. In this formulation, the AST is not merely a
parsing artifact, but the mechanism that enables unambiguous reference to semantic
program elements.

Rather than interacting with repositories through anonymous text spans, 
\textsc{\system} agents act through a small set of structure-aware primitives that operate
directly on named AST entities. Concretely, the action space consists of two
primitive operations: \texttt{readCode} and \texttt{editCode}. Each action is
parameterized by a selector that identifies a target program entity, allowing the
agent to specify \emph{what} it intends to inspect or modify without committing to
\emph{how} that change is realized in source text.

A \texttt{readCode} action retrieves code for a selected AST node,
returning complete syntactic units (e.g., functions or classes) or compact
structural summaries. An \texttt{editCode} action performs AST-level
transformations---such as insertion, replacement, or removal---to a selected
entity, producing a modified AST that is syntactically valid by construction.

Each \texttt{editCode} action transforms the current AST into a
new, syntactically valid AST. Consequently, a multi-step repository-level editing
process can be viewed as a trajectory of structured actions over successive AST
states. This formulation yields explicit and analyzable action traces, enabling
fine-grained analysis of agent behavior beyond final patch correctness.

By grounding all interactions in named program entities, \textsc{\system}
explicitly separates semantic intent from textual realization. This design avoids
brittle dependencies on line numbers or string matching, ensures that edits respect
syntactic boundaries, and aligns the agent’s action space with the abstractions
used by human developers.

\subsection{Structure-Aware Tools}
\label{sec:tools}

We instantiate this action space through two core tools: a structure-aware read operation (\texttt{readCode}) and a structure-aware edit operation (\texttt{editCode}). Together, these tools define the primitive interactions available to an agent. Algorithms~\ref{alg:code_read} and~\ref{alg:code_write} formalize these operations.

\begin{algorithm}[t]
\small
\caption{\textsc{readCode}: Structure-Aware Code Retrieval}
\label{alg:code_read}
\begin{algorithmic}[1]
\REQUIRE File path $p$, file size threshold $\tau$, selector $\sigma$ (optional), line range $(l_s, l_e)$ (optional)
\ENSURE Code content or structural summary
\IF{$l_s, l_e$ specified}
    \RETURN $\textsc{ExtractLines}(p, l_s, l_e)$
\ENDIF
\IF{$\sigma = \emptyset$ \textbf{and} $|p| < \tau$}
    \RETURN the full file content of $p$
\ENDIF
\STATE $\mathcal{T} \gets \textsc{ParseAST}(p)$ \COMMENT{Parse file into AST}
\STATE $\mathcal{E} \gets \textsc{ExtractSignatures}(\mathcal{T})$ \COMMENT{Extract functions, classes}
\IF{$\sigma = \emptyset$}
    \RETURN $\textsc{FormatSignatures}(\mathcal{E})$ \COMMENT{Structural summary}
\ELSE
    \STATE $\mathcal{M} \gets \textsc{FuzzyMatch}(\mathcal{E}, \sigma)$ \COMMENT{Selector matching}
    \RETURN $\{e.\text{impl} : e \in \mathcal{M}\}$
\COMMENT{$e.\text{impl}$ denotes the full source span of entity $e$ (its AST subtree rendered as code)}
\ENDIF
\end{algorithmic}
\end{algorithm}

\begin{algorithm}[t]
\small
\caption{\textsc{editCode}: Structure-Aware Code Modification}
\label{alg:code_write}
\begin{algorithmic}[1]
\REQUIRE File path $p$, operation $\omega \in \{\texttt{insert}, \texttt{replace}, \texttt{removal}\}$, selector $\sigma$, replacement code $r$ %\dm{change delete to removal to keep consistent with Fig 1.}
\ENSURE Modified file with syntactic validity guarantee
\STATE $\mathcal{T} \gets \textsc{GetOrParseAST}(p)$ \COMMENT{Reuse cached AST if available}
\STATE $n \gets \textsc{FindNode}(\mathcal{T}, \sigma)$ \COMMENT{Locate target AST node}
\IF{$n = \emptyset$}
    \RETURN \textsc{Error}(``Selector not found'')
\ENDIF
\STATE $\delta \gets \textsc{GetIndentation}(n)$ \COMMENT{Preserve formatting}
\STATE $r' \gets \textsc{ApplyIndentation}(r, \delta)$
\IF{$\omega = \texttt{insert}$}
    \STATE $\mathcal{T}' \gets \textsc{InsertAfter}(\mathcal{T}, n, r')$
\ELSIF{$\omega = \texttt{replace}$}
    \STATE $\mathcal{T}' \gets \textsc{ReplaceNode}(\mathcal{T}, n, r')$
\ELSIF{$\omega = \texttt{removal}$}
    \STATE $\mathcal{T}' \gets \textsc{RemoveNode}(\mathcal{T}, n)$
\ENDIF
\IF{$\textsc{HasSyntaxError}(\mathcal{T}')$}
    \RETURN \textsc{Error}(``Invalid syntax'') \COMMENT{Reject malformed edits}
\ENDIF
\STATE \textsc{WriteFile}(p, $\mathcal{T}'$)
\RETURN \textsc{Success}
\end{algorithmic}
\end{algorithm}

\paragraph{\texttt{readCode}}
This tool provides structure-aware code retrieval by exposing named program entities
as the primary units of access (Algorithm~\ref{alg:code_read}). The tool supports a
coarse-to-fine workflow with three common modes.

\textbf{(1) Repository browsing (directory input).}
When the input path $p$ is a directory, \texttt{readCode} returns the list of files under $p$
(optionally filtered to source files). This enables agents to navigate the repository layout
before selecting a file to inspect.

\textbf{(2) File summarization (file input, no selector).}
When $p$ is a file and no selector $\sigma$ is provided, the tool adaptively returns either
(i) the full file content if the file is small (below a size threshold $\tau$, e.g., 10K characters),
or (ii) a compact structural summary if the file is large. The summary consists of signatures of
top-level entities (e.g., classes and functions) and their scoped names, allowing the agent to
identify relevant program entities without loading the entire file.

\textbf{(3) Entity retrieval (file + selector).}
When $p$ is a file and a selector $\sigma$ is provided, \texttt{readCode} resolves $\sigma$
to one or more program entities extracted from the file’s AST and returns the complete
implementation of each matched entity (i.e., the AST subtree rendered back into source code).
Selectors can be \emph{unscoped} (e.g., \texttt{load}) or \emph{scoped} (e.g.,
\texttt{User.load}) to restrict matches to methods within a specific class.
Selector resolution uses deterministic name-based matching; for example, \texttt{guf} can match
\texttt{get\_user\_file}, and \texttt{User.load} matches method \texttt{load} in class \texttt{User}.
This matching is deterministic and does not involve learned components.

\texttt{readCode} encourages agents to first discover relevant entities 
via directory browsing and structural summaries, then selectively 
retrieve only the implementations needed for reasoning. Unlike 
line-range reading, selector-based retrieval returns complete 
syntactic units, reducing irrelevant context and avoiding brittle 
dependence on line numbers.

\paragraph{\texttt{editCode}}
The tool performs structure-aware  modification by applying
AST-grounded transformations to named program entities (Algorithm~\ref{alg:code_write}).
Each edit is specified by an operation type $\omega$ (insertion, replacement, or removal)
and a selector $\sigma$ that identifies the target entity in the AST.

Given a selector, \texttt{editCode} locates its associated AST node and applies the
requested transformation within the node’s syntactic scope. The tool automatically
preserves formatting by computing the local indentation context and validating the
modified AST before committing the change. Edits that would introduce syntax errors are
rejected, ensuring post-edit syntactic validity via AST validation.

By exposing edits as atomic operations over named entities, \texttt{editCode} enables
agents to perform targeted and interpretable modifications such as adding a new method,
deleting an obsolete function, or replacing the implementation of an existing routine.
This design separates semantic intent from textual realization: the agent specifies
\emph{what} to change via an entity-level selector, while the tool determines \emph{how} to apply it in the source text.

This syntactic validity guarantee distinguishes \textsc{\system} from text-based editing
approaches, where edits are applied via line numbers or string matching and malformed
changes can silently corrupt the codebase. Each \texttt{editCode} invocation produces an
explicit and traceable state transition, enabling fine-grained analysis of agent
trajectories beyond final patch correctness.

\paragraph{Agent Interface and Integration}
\texttt{readCode} and \texttt{editCode} are exposed through a
standardized tool interface, allowing them to be invoked by arbitrary LLM-based
agents. In particular, the interface is implemented using the Model Context
Protocol (MCP), which is supported by most existing agent frameworks. As a result,
\textsc{\system} can be integrated into off-the-shelf agents without modifying
their planning or execution logic, enabling the structured action space to be
adopted independently of agent-specific infrastructure.

% By exposing edits as structured operations over named entities, \texttt{editCode} enables agents to perform targeted, interpretable modifications such as adding a new method, deleting an obsolete function, or replacing the implementation of an existing routine. This syntactic validity guarantee distinguishes \textsc{\system} from text-based editing approaches, where malformed edits can silently corrupt the codebase. Edits are executed as atomic transformations, producing explicit and traceable action trajectories that can be analyzed independently of the underlying model.

% We provide a detailed comparison of AST-based and text-based tool trajectories on a concrete SWE-bench instance in Appendix~\ref{appendix:tool-comparison}, illustrating how structure-aware operations reduce navigation overhead and avoid common failure modes such as whitespace mismatch errors.

\section{Experiments}
\label{sec:experiments}

\subsection{Tasks and Datasets}

We evaluate \textsc{\system} on repository-level software engineering benchmarks requiring agents to perform multi-step code understanding and modifications across multiple files and program entities. 
% Unlike single-function completion tasks, these benchmarks demand reasoning about existing code structure, identifying semantically relevant program elements, and applying targeted edits that preserve program correctness.

\paragraph{SWE-Bench Verified.} This benchmark consists of 500 real-world Python GitHub issues paired with failing tests that specify the desired behavior~\cite{jimenez2024swebench} . Solving a task requires locating the relevant code regions, understanding the underlying bug or feature request, and modifying one or more program entities to satisfy the test cases. SWE-Bench-Verified has emerged as a standard benchmark for evaluating repository-level program repair systems, where success is primarily measured by the correctness of applied code edits. These tasks naturally stress an agent’s ability to navigate and manipulate codebases through structured interactions rather than ad hoc text edits. 
% \joe{add dataset stats}

\paragraph{CodeAssistBench Verified.} It consists of 135 multi-turn programming assistance tasks across 7 programming languages derived from real-world GitHub issues that involve clarification, code exploration, and iterative refinement~\cite{kim2025codeassistbench}.
Tasks in CodeAssistBench frequently require agents to inspect multiple functions or classes, reason about their relationships, and apply localized changes over several interaction steps. Unlike SWE-Bench, CodeAssistBench is not exclusively focused on producing a final patch, but instead evaluates an agent’s ability to support interactive and exploratory programming workflows.
% \joe{add dataset stats}

% \paragraph{Rationale for Multiple Benchmarks.}
% We include both SWE-Bench and CodeAssistBench to evaluate the generality of structure-aware action spaces across different task formulations and agent behaviors. SWE-Bench emphasizes edit-centric program repair and aligns closely with SWE-Agent-style workflows, while CodeAssistBench captures more interactive settings in which agents must balance code exploration, partial edits, and user guidance. Evaluating across these complementary benchmarks allows us to demonstrate that the benefits of \textsc{\system} are not limited to a single agent design or benchmark, but instead arise from exposing code structure as a programmable action space that generalizes across both edit-focused and exploratory programming assistance scenarios. \joe{just merge back to tasks and datasets section 4.1}

% \paragraph{Task Characteristics.}
% Across both benchmarks, successful task completion typically involves (1) identifying a small subset of semantically relevant program entities within a large repository, (2) selectively reading their implementations, and (3) applying targeted modifications that respect syntactic and structural constraints. These shared characteristics align closely with the programmable AST action space exposed by \textsc{\system}, making these benchmarks well suited for evaluating the impact of structure-aware agent interactions. \joe{just merge back to tasks and datasets section 4.1}

\subsection{Baselines}

We compare \textsc{\system} with representative baselines that differ in how agents interact with code repositories. 
% To isolate the impact of the action space, all methods use identical language models, task prompts, and execution budgets; the only difference lies in the interfaces exposed to the agent for reading and modifying code.

\begin{table*}[t]
\centering
\footnotesize
\setlength{\tabcolsep}{4pt}
\begin{tabular}{llrrrrr}
\toprule
\textbf{Model} & \textbf{Interface}
& \textbf{Pass@1 (\%) $\uparrow$}
& \textbf{Input Tokens}
& \textbf{Output Tokens}
& \textbf{LLM Calls $\downarrow$}
& \textbf{Cost (\$) $\downarrow$} \\
\midrule

\multirow{2}{*}{GPT-5}
& Baseline
& 66.0
& 452.7M
& 0.81M
& 16{,}436
& 574.0 \\
& \textbf{CodeStruct}
& \textbf{67.2} \posc{(+1.2)}
& \textbf{366.3M} \posc{(-19.1\%)}
& \textbf{0.44M} \posc{(-45.7\%)}
& \textbf{16{,}307} \posc{(-0.8\%)}
& \textbf{462.2} \posc{(-19.5\%)} \\
\midrule

\multirow{2}{*}{GPT-5-mini}
& Baseline
& 60.4
& 593.7M
& 1.27M
& 18{,}560
& 151.0 \\
& \textbf{CodeStruct}
& \textbf{62.0} \posc{(+1.6)}
& \textbf{404.5M} \posc{(-31.9\%)}
& \textbf{0.35M} \posc{(-72.4\%)}
& \textbf{14{,}811} \posc{(-20.2\%)}
& \textbf{101.8} \posc{(-32.6\%)} \\
\midrule

\multirow{2}{*}{GPT-5-nano}
& Baseline
& 19.6
& 808.0M
& 0.86M
& 24{,}037
& \textbf{40.7} \\
& \textbf{CodeStruct}
& \textbf{40.4} \textbf{\posc{(+20.8)}}
& 1{,}137.4M \negc{(+40.8\%)}
& 0.95M \negc{(+10.5\%)}
& 27{,}278 \negc{(+13.5\%)}
& 57.3 \negc{(+40.8\%)} \\
\midrule

\multirow{2}{*}{Qwen3-Coder}
& Baseline
& 61.2
& 805.8M
& 1.50M
& 26{,}961
& 365.3 \\
& \textbf{CodeStruct}
& \textbf{66.2} \posc{(+5.0)}
& \textbf{705.3M} \posc{(-12.5\%)}
& \textbf{2.17M} \negc{(+44.6\%)}
& \textbf{32{,}346} \negc{(+20.0\%)}
& \textbf{321.3} \posc{(-12.1\%)} \\

\midrule

\multirow{2}{*}{Qwen3-32B}
& Baseline
& 14.8
& 366.0M
& 1.23M
& 25{,}543
& 55.6 \\
& \textbf{CodeStruct}
& \textbf{16.0} \posc{(+1.2)}
& \textbf{302.1M} \posc{(-17.5\%)}
& \textbf{1.03M} \posc{(-16.3\%)}
& \textbf{24{,}653} \posc{(-3.5\%)}
& \textbf{45.9} \posc{(-17.4\%)} \\
\midrule
\multirow{2}{*}{Qwen3-8B}
& Baseline
& 13.2
& 84.0M
& 0.11M
& 8{,}833
& 2.36 \\
& \textbf{CodeStruct}
& 13.0 \negc{(-0.2)}
& \textbf{51.8M} \posc{(-38.3\%)}
& \textbf{0.08M} \posc{(-27.3\%)}
& \textbf{8{,}313} \posc{(-5.9\%)}
& \textbf{1.46} \posc{(-38.1\%)} \\
\bottomrule

\end{tabular}

\caption{
Main results on SWE-Bench Verified.
Agents follow SWE-Agent-style workflows---iterative loops that alternate between reading repository files and applying code edits via tool calls.
We compare text-based interaction (\emph{Baseline}) against \textsc{\system}.
Inline percentages indicate relative change compared to \emph{Baseline}
(percentage points for Pass@1; relative \% for tokens, calls, and cost).
Green denotes improvement; red denotes regression.
}
\label{tab:swebench-main}
\end{table*}

\paragraph{Baseline Text-Based Agents.}
We compare against SWE-Agent~\cite{yang2024sweagent} and 
OpenHands~\cite{wang2024openhands}, which represent the dominant 
text-based interaction paradigm. Both systems retrieve code by 
reading entire files or line ranges and apply modifications via 
string replacement. To ensure a fair comparison, we enable 
SWE-Agent's repository map feature, which exposes file structure 
and function signatures as navigation hints. Despite this structural 
guidance, all read and edit operations remain text-based: the agent 
must specify line numbers or match exact strings to retrieve or 
modify code. This configuration represents the strongest reasonable 
baseline within the text-based paradigm.

We note that structural summaries such as repository maps, symbol indices,
and dependency graphs operate at a different and complementary level from
\textsc{\system}: they primarily support \emph{navigation and planning}
(deciding \emph{where} to look), while \textsc{\system} targets the
\emph{action interface} (redefining \emph{how} agents read and edit code
once a target is identified).
Since our baseline already includes such structural summaries, the
comparison isolates the effect of changing the read/edit action space
while holding navigation aids constant.

\paragraph{\system Agents.}
In contrast, agents using \system interact with repositories through structure-aware \texttt{readCode} and \texttt{editCode} tools.
\texttt{readCode} is scoped to named functions, classes, or methods, while \texttt{editCode} is executed as AST-level transformations that preserve syntactic validity. This interface enables agents to directly operate over program structure, rather than reasoning indirectly through unstructured text.

% \paragraph{Agent Variants.}
% To evaluate robustness across agent designs, we instantiate both text-based and structure-aware interfaces using multiple agent implementations. These include edit-centric agents aligned with SWE-Agent-style workflows, as well as more general-purpose agents such as OpenHands. Crucially, each agent is evaluated under both interface settings, ensuring that observed performance differences can be attributed to the underlying action space rather than agent-specific planning heuristics or control logic.
% \joe{maybe we can drop it for saving space}

% \vk{Ideally, we should have results with open-hands and mini-swe agent as these are common baselines. Readers don't know what is Strands or Strands-style agents}
% To evaluate robustness across agent designs, we instantiate both text-based and structure-aware interfaces using multiple agent implementations, including edit-centric agents aligned with SWE-Agent-style workflows and more general agents Open Hands. This allows us to assess whether observed improvements stem from the underlying action space rather than agent-specific heuristics or control logic.

\subsection{Experimental Setup}

\paragraph{Models.}
We evaluate all methods using a diverse set of large language models that span both proprietary and open-weight families, including GPT-5~\cite{openai_chatgpt_2024}, GPT-5-mini, GPT-5-nano, Qwen3-480B-A30B-Coder~\cite{yang2025qwen3}, Qwen3-32B, and Qwen3-8B. Unless otherwise specified, all models are used with default decoding parameters. No method is provided with model-specific fine-tuning or task-specific adaptations.

% \paragraph{Agent Configuration.}
% \joe{should merge to baselines and save space}
% All agents share the same underlying SWE-Agent\footnote{All variants are based on the same \href{https://github.com/SWE-agent/SWE-agent/commit/1d3cfb798a2c257a0fc6094f1d35ef084d9919e1}{SWE-Agent commit}.
%  } control logic
% with identical prompts, step limits, termination conditions, and error handling.
% The \emph{Interface} column in Table~\ref{tab:swebench-main}
% indicates which read and write action interface is exposed to the agent
% (\emph{Original} text-based tools or \textsc{\system});
% all other components remain identical.

% \paragraph{Agent Configuration.}
% \vk{ Interface col in the table in undefined. Which version of the agent are we using here}
% All agents follow the same high-level control logic for iterative repository interaction, including step limits, termination conditions, and error handling. For each task, an agent iteratively selects read or write actions until either the task is solved, the execution budget is exhausted, or no further progress can be made.

\paragraph{Prompts and Budgets.}
To isolate the effect of the action interface, we do not modify the system or task prompts between the baseline and CODESTRUCT, and use the default prompts provided by the underlying agent frameworks. We fix a maximum interaction budget for each task to \$5 for large models (GPT-5 and Qwen3-480B-A30B-Coder), \$3 for mid-size models (GPT-5-mini and Qwen3-32B), and \$1 for small-size models (GPT-5-nano and Qwen3-8B).

% \paragraph{Action Space Variants.}
% To isolate the impact of the interaction interface, the only difference between methods lies in the action space exposed to the agent. Text-based baselines operate over file- or line-level reads and string-based edits, while \textsc{\system} exposes structure-aware read and write actions over semantic program entities. No additional supervision, oracle information, or task-specific heuristics are provided to any method. \joe{merge to baselines the same as agent config section}

% \paragraph{Evaluation Protocol.}
% For SWE-Bench, a task is considered successful if all associated tests pass after the agent completes its interactions with the repository. For CodeAssistBench, we follow the benchmark’s standard evaluation protocol, which assesses task completion based on the correctness and usefulness of the agent’s final outputs and applied modifications. All results are averaged over the same task splits for each benchmark. \joe{drop it and make sure we mention it in the dataset section}

\begin{table*}[t]
\centering
\footnotesize
\setlength{\tabcolsep}{4.2pt}
\begin{tabular}{l l r r r r r}
\toprule
\textbf{Model} & \textbf{Interface}
& \textbf{Accuracy (\%) $\uparrow$}
& \textbf{Input Tokens}
& \textbf{Output Tokens}
& \textbf{LLM Calls $\downarrow$}
& \textbf{Cost (\$) $\downarrow$} \\
\midrule

\multirow{2}{*}{GPT-5}
& Baseline
& 53.3
& 143.9M
& 1.27M
& 566
& \$19.57 \\
& \textbf{CodeStruct}
& \textbf{54.1} \posc{(+0.8)}
& \textbf{122.1M} \posc{(-15.1\%)}
& \textbf{1.18M} \posc{(-7.0\%)}
& \textbf{550} \posc{(-2.8\%)}
& \textbf{\$16.74} \posc{(-14.5\%)} \\
\midrule

\multirow{2}{*}{GPT-5-mini}
& Baseline
& 51.1
& 125.3M
& 1.26M
& 648
& \$3.45 \\
& \textbf{CodeStruct}
& \textbf{51.9} \posc{(+0.8)}
& \textbf{83.1M} \posc{(-33.7\%)}
& \textbf{0.89M} \posc{(-28.7\%)}
& \textbf{630} \posc{(-2.8\%)}
& \textbf{\$2.30} \posc{(-33.3\%)} \\
\midrule

\multirow{2}{*}{GPT-5-nano}
& Baseline
& 46.7
& 56.0M
& 1.25M
& 742
& \$0.34 \\
& \textbf{CodeStruct}
& \textbf{48.1} \posc{(+1.4)}
& \textbf{52.8M} \posc{(-5.7\%)}
& \textbf{1.05M} \posc{(-15.6\%)}
& \textbf{652} \posc{(-12.1\%)}
& \textbf{\$0.32} \posc{(-5.9\%)} \\
\midrule

\multirow{2}{*}{Qwen3-Coder}
& Baseline
& 31.1
& 829.1M
& 6.95M
& 796
& \$385.61 \\
& \textbf{CodeStruct}
& \textbf{31.9} \posc{(+0.8)}
& \textbf{779.2M} \posc{(-6.0\%)}
& 7.00M \negc{(+0.7\%)}
& 826 \negc{(+3.8\%)}
& \textbf{\$363.24} \posc{(-5.8\%)} \\
\midrule

\multirow{2}{*}{Qwen3-32B}
& Baseline
& 15.6
& 110.5M
& 2.18M
& 858
& \$29.01 \\
& \textbf{CodeStruct}
& \textbf{20.0} \posc{(+4.4)}
& 142.4M \negc{(+28.9\%)}
& \textbf{1.25M} \posc{(-42.7\%)}
& \textbf{776} \posc{(-9.6\%)}
& \$38.71 \negc{(+23.7\%)} \\

\midrule

\multirow{2}{*}{Qwen3-8B}
& Baseline
& 13.3
& 0.63M
& 0.21M
& 984
& \$0.05 \\
& CodeStruct
& 14.1 \posc{(+0.8)}
& \textbf{0.55M} \posc{(-12.7\%)}
& \textbf{0.17M} \posc{(-21.7\%)}
& \textbf{944} \posc{(-4.1\%)}
& \textbf{\$0.04} \posc{(-17.6\%)} \\

\bottomrule
\end{tabular}
\caption{
CodeAssistBench results comparing text-based interaction (\emph{Baseline})
against \textsc{CodeStruct}.
Inline values show relative change over \emph{Baseline}
(percentage points for Accuracy; relative \% for tokens, calls, and cost).
}
\label{tab:codeassistbench-main}
\vspace{-15pt}
\end{table*}

\subsection{Results}
\subsubsection{SWE-Bench Verified Results}
\label{sec:results-swebench}

Table~\ref{tab:swebench-main} reports the main results of integrating
\textsc{\system} into SWE-Agent-style workflows on SWE-Bench Verified.
Beyond accuracy, \textsc{\system} substantially improves interaction efficiency.
Across most models, token consumption is reduced by 12--38\%,
reflecting fewer large file reads and more targeted access to relevant program elements.
Output token usage also decreases consistently for most models,
indicating that structure-aware actions reduce redundant rewrite attempts
and patch churn, rather than merely shifting cost from reads to edits.
These improvements are accompanied by reductions in the number of API calls,
suggesting that agents converge to correct solutions in fewer interaction steps.

Reductions in token usage directly translate into lower inference cost, where
\textsc{\system} reduces the total inference cost by 19.5\% for GPT-5,
32.6\% for GPT-5-mini, and 17.4\% for Qwen3-32B, while simultaneously improving Pass@1.

\paragraph{Addressing Motivating Limitations.}
Aggregate improvements in Table~\ref{tab:swebench-main} reflect
\textsc{\system}'s impact each motivating limitation in the Introduction. We explicitly map each to its empirical proxy:

\textbf{Irrelevant context} \textrightarrow\ \textbf{Input tokens.}
Selector-based retrieval returns only the targeted syntactic unit rather
than entire files, reducing input tokens by 12–38\% across most models
(Table~\ref{tab:swebench-main}); removing \texttt{readCode} reverses this trend,
increasing input tokens by +41\% for Qwen3-32B
(Table~\ref{tab:ablation}).
On \texttt{django\_\_django-11211}, the text-based agent reads
\textasciitilde300 lines while our \textsc{\system} agent reads \textasciitilde50
(Figure~\ref{fig:overview}).

\textbf{Wasteful exploration} \textrightarrow\ \textbf{Interaction steps
and LLM calls.} On the same instance, localization drops from 21 steps
to 2 (Table~\ref{tab:django-12965-comparison};
Appendix~\ref{appendix:tool-comparison}).
Across all instances, LLM calls decrease by up to 20.2\%
(GPT-5-mini). Without \texttt{readCode}, \texttt{str\_replace} calls
increase 7.8$\times$ on regressed instances
(Appendix~\ref{sec:ablation}).

\textbf{Brittle string-matching edits} \textrightarrow\ \textbf{Tool-level
errors and empty patches.} For capable models, edit errors per instance
decrease by 76--88\% (Table~\ref{tab:error-analysis}).
Empty-patch failures drop from 233 to 36 for GPT-5-nano
(-84.5\%; Appendix~\ref{appendix:empty-patch}).

\textbf{Redundant code regeneration} \textrightarrow\ \textbf{Output
tokens.} Structure-aware edits specify only the entity name and new
content, avoiding verbatim reproduction of surrounding code. Output
tokens decrease by 45.7\% for GPT-5 and 72.4\% for GPT-5-mini.

Two models exhibit increased token usage under \textsc{\system}.
For GPT-5-nano, \textsc{\system} achieves substantially higher
accuracy at the expense of increased computation.
This reflects a natural tradeoff under limited model capacity:
structure-aware actions encourage deeper exploration and sustained interaction
rather than early termination, increasing compute while improving solution quality.
For Qwen3-Coder (480B), although input tokens decrease by 12.5\%,
output tokens increase by 45\% and LLM calls rise by 20\%.
We therefore do not claim efficiency improvements for this model.
Instead, the benefit is accuracy: +5.0pp in Pass@1 at a 12.1\% lower
total cost, since input tokens dominate the cost.
This suggests that \textsc{\system} leads the model to spend more effort
on intermediate reasoning and problem localization rather than on repeated
text-based editing attempts. This indicates that \textsc{\system}'s benefits are not limited to
uniform efficiency gains, but can also arise from improving the quality
of agent exploration under a fixed interaction budget.

\paragraph{Code Editing Error Analysis.} To understand how structured interfaces affect operational reliability, we analyze tool-level error patterns—failed edit operations—across all agent trajectories (see Appendix~\ref{appendix:error-analysis} for error analysis and Appendix~\ref{appendix:empty-patch} for empty-patch analysis). Table~\ref{tab:error-analysis} reveals a clear capability-dependent effect. For higher-capacity models (GPT-5, GPT-5-mini, Qwen3-Coder), \textsc{\system} reduces errors per instance by 76–88\%, indicating that AST-based operations are substantially more reliable than text-based string matching.

For Qwen3-8B and Qwen3-32B, \textsc{\system} also substantially reduces tool-level errors, confirming that structured actions mitigate interface brittleness even for weaker models; however, error counts remain high in absolute terms (10–14 per instance), and accuracy gains are limited.

In contrast, GPT-5-nano exhibits a 20\% increase in tool-level errors despite a 20.8pp accuracy gain, reflecting a redistribution rather than a reduction of failures: structured navigation enables correct localization and more edit attempts while dramatically reducing early agent terminations (empty patches drop from 233 to 36).

\paragraph{Ablation Study.}
To isolate the contribution of each component, we evaluate \textsc{\system} with either \texttt{readCode} or \texttt{editCode} removed (Table~\ref{tab:ablation}; detailed analysis in Appendix~\ref{sec:ablation}). Both primitives contribute to effectiveness, but in complementary ways. Removing \texttt{readCode} causes the largest accuracy degradation ($-7.8$ Pass@1 for Qwen3-32B, $-5.2$ for GPT-5-mini), accompanied by substantially higher token usage and more LLM calls, indicating that without structured navigation, agents resort to inefficient trial-and-error exploration. Notably, Qwen3-32B without \texttt{readCode} underperforms even the text-based baseline (8.2\% vs.\ 14.8\%). This occurs because the hybrid configuration creates a mismatch: the agent's prompts and planning still expect structured navigation (e.g., selector-based retrieval), but without \texttt{readCode} it must fall back to text-based exploration, leading to less coherent search strategies than the fully text-based baseline where all tools are mutually consistent. Analysis of regressed instances confirms this: \texttt{str\_replace} calls increase by 7.8$\times$ and the dominant failure mode shifts from incorrect patches to budget exhaustion (Appendix~\ref{sec:ablation}).
Removing \texttt{editCode} yields smaller accuracy drops but disproportionate cost penalties: GPT-5-mini incurs a 38.7\% cost increase for only 1.4pp less accuracy, as agents fall back to brittle string-based edits requiring more validation cycles. These results confirm that \texttt{readCode} and \texttt{editCode} serve complementary roles: structured navigation minimizes exploration cost while structured editing minimizes transformation cost.

\paragraph{Qualitative Analysis.}
To illustrate these efficiency gains concretely, we compare AST-based and text-based agent trajectories on representative SWE-Bench instances (Appendix~\ref{appendix:tool-comparison}). On \texttt{django\_\_django-11211}, the text-based agent spends 21 steps navigating via \texttt{grep}, \texttt{sed}, and line-range reads before locating the target method, while \system achieves the same localization in 2 steps using selector-based access. This reduces total steps from 54 to 24—a 55\% reduction—demonstrating how structure-aware navigation eliminates the trial-and-error exploration common in text-based approaches.

Overall, these results demonstrate that exposing the codebase as a structured
action space improves both the effectiveness and efficiency of SWE-Agent on SWE-Bench,
while also reducing interaction-level failure modes,
without requiring model-specific tuning or changes to the agent's
decision-making logic.

\subsubsection{Results on CodeAssistBench}
\label{sec:cab-results}

We further evaluate \textsc{\system} on \textbf{CodeAssistBench (CAB)},
a benchmark designed to assess multi-turn, interactive code-assistance scenarios.
Unlike SWE-Bench, CAB emphasizes conversational problem solving and
incremental tool use rather than single-shot patch generation.
We use OpenHands, the default agent framework provided by CodeAssistBench,
and added read and edit operations with \textsc{CodeStruct}.

Table~\ref{tab:codeassistbench-main} compares the baseline text-based
interaction interface against \textsc{CodeStruct} with structure-aware
AST read and edit actions across six models.
\textsc{CodeStruct} consistently improves accuracy while reducing token usage and cost for most models.
Among GPT-family models, \textsc{CodeStruct} achieves accuracy gains of
$+0.8$ to $+1.4$ percentage points while reducing resource consumption.
GPT-5-mini shows the largest efficiency improvement, with input tokens reduced by $33.7\%$
and cost reduced by $33.3\%$.
GPT-5-nano achieves the highest accuracy gain ($+1.4$) alongside a $12.1\%$ reduction in LLM calls.
For Qwen models, results are more nuanced.
Qwen3-32B shows the largest accuracy improvement ($+4.4$), though at the cost of increased input tokens ($+28.9\%$).
This shows that for weaker models, \textsc{CodeStruct}'s structured actions enable more thorough exploration that improves solution quality.
In contrast, Qwen3-Coder and Qwen3-8B show modest accuracy gains with improved efficiency.

Overall, these results demonstrate that structured program interactions
generalize beyond patch-based benchmarks to interactive code-assistance workflows,
where efficient exploration and precise context selection are critical.

\section{Conclusion}

We introduced \textsc{\system}, a structure-aware interface that exposes a codebase as a programmable action space for LLM-based agents. Rather than interacting with repositories through unstructured text spans, agents operate over named program entities derived from the AST, enabling semantically grounded reads and syntax-preserving edits. This design directly addresses the abstraction mismatch in existing code agents, which treat programs as flat text despite their inherently structured nature, and provides a principled alternative for repository-level code reasoning and modification. Across SWE-Bench Verified and CodeAssistBench Verified, we showed that replacing text-based read and edit operations with structure-aware actions improves both effectiveness and efficiency. 
\textsc{\system} reduces unnecessary context retrieval, lowers inference cost, and mitigates brittle string-based failures, with particularly large gains for models whose failures stem from 
text-interface brittleness rather than reasoning limitations.

\section*{Limitations}
\paragraph{File-Level AST Scope.}
\textsc{\system} currently operates on per-file ASTs and does not explicitly model cross-file dependencies such as inheritance hierarchies or inter-module call graphs. We note that text-based baselines also operate at the file level with no cross-file semantic modeling, so this is a shared limitation of current agent interfaces rather than one specific to \textsc{\system}. Our strongest baseline already includes SWE-Agent's repository map, which provides file-level structural summaries and symbol indices as navigation hints; \textsc{\system} is complementary to such mechanisms, changing \emph{how} agents read and edit code once a target file is identified, rather than \emph{where} they look. Incorporating cross-file structure (e.g., inheritance hierarchies, call graphs) is a promising direction, though it involves a practical trade-off: file-level AST parsing is stateless and completes in milliseconds, while cross-file analysis requires whole-repository indexing that must be updated after every edit.

\paragraph{Language Coverage.}
Our bug-fixing evaluation on SWE-Bench Verified focuses on Python, though CodeAssistBench provides additional coverage across seven languages. Extending the evaluation to other languages with different syntactic characteristics (e.g., statically-typed languages with complex generics) is a natural direction for future work.

\paragraph{AST Parsing Overhead.}
AST construction introduces additional tool execution time relative to raw text operations. However, this overhead is negligible compared to LLM inference latency. Using tree-sitter for local parsing, median execution times are 146--171ms for \texttt{readCode} and 189--212ms for \texttt{editCode}, while median LLM call latency ranges from 4--12 seconds. Parsed ASTs are cached via \textsc{GetOrParseAST} (Algorithm~\ref{alg:code_write}, line~1) to avoid redundant parsing. End-to-end, all AST operations consumed 35.7 minutes across 500 SWE-Bench runs with GPT-5 (75.95 compute hours total), accounting for less than 0.8\% of total runtime---substantially outweighed by the 12--38\% reduction in LLM tokens and inference calls.

\paragraph{Robustness to Syntax Errors.}
\system requires syntactically valid source files for AST parsing. While our \texttt{editCode} tool includes syntax validation to reject malformed edits and preserve repository integrity (Algorithm~\ref{alg:code_write}), files that are already syntactically invalid prior to agent interaction cannot benefit from structure-aware operations. In practice, committed code in software repositories is overwhelmingly syntactically valid, limiting the impact of this constraint.

\paragraph{Task Coverage.}
We evaluate on two core repository-level coding tasks: bug fixing (SWE-Bench Verified) and interactive code assistance (CodeAssistBench). Evaluating structure-aware action spaces on additional tasks such as code review and test generation remains future work.

\bibliography{custom}

\newpage
\appendix
\onecolumn

\section{Appendix: Detailed Tool Comparison Example}
\label{appendix:tool-comparison}

We illustrate the difference between AST-based and text-based code navigation tools using the SWE-bench instance \texttt{django\_\_django-11211}. The task requires fixing Django's \texttt{GenericForeignKey.get\_prefetch\_queryset} method to correctly handle primary key type conversion.

\subsection{Problem Description}

When prefetching \texttt{GenericForeignKey} relations, the lookup fails because the stored \texttt{object\_id} (a string) is compared against the model's primary key (an integer) without type conversion. The fix requires modifying the \texttt{gfk\_key} function inside \texttt{get\_prefetch\_queryset} to use \texttt{to\_python()} instead of \texttt{get\_prep\_value()}.

\begin{figure*}[t]
    \centering
    \includegraphics[width=\textwidth]{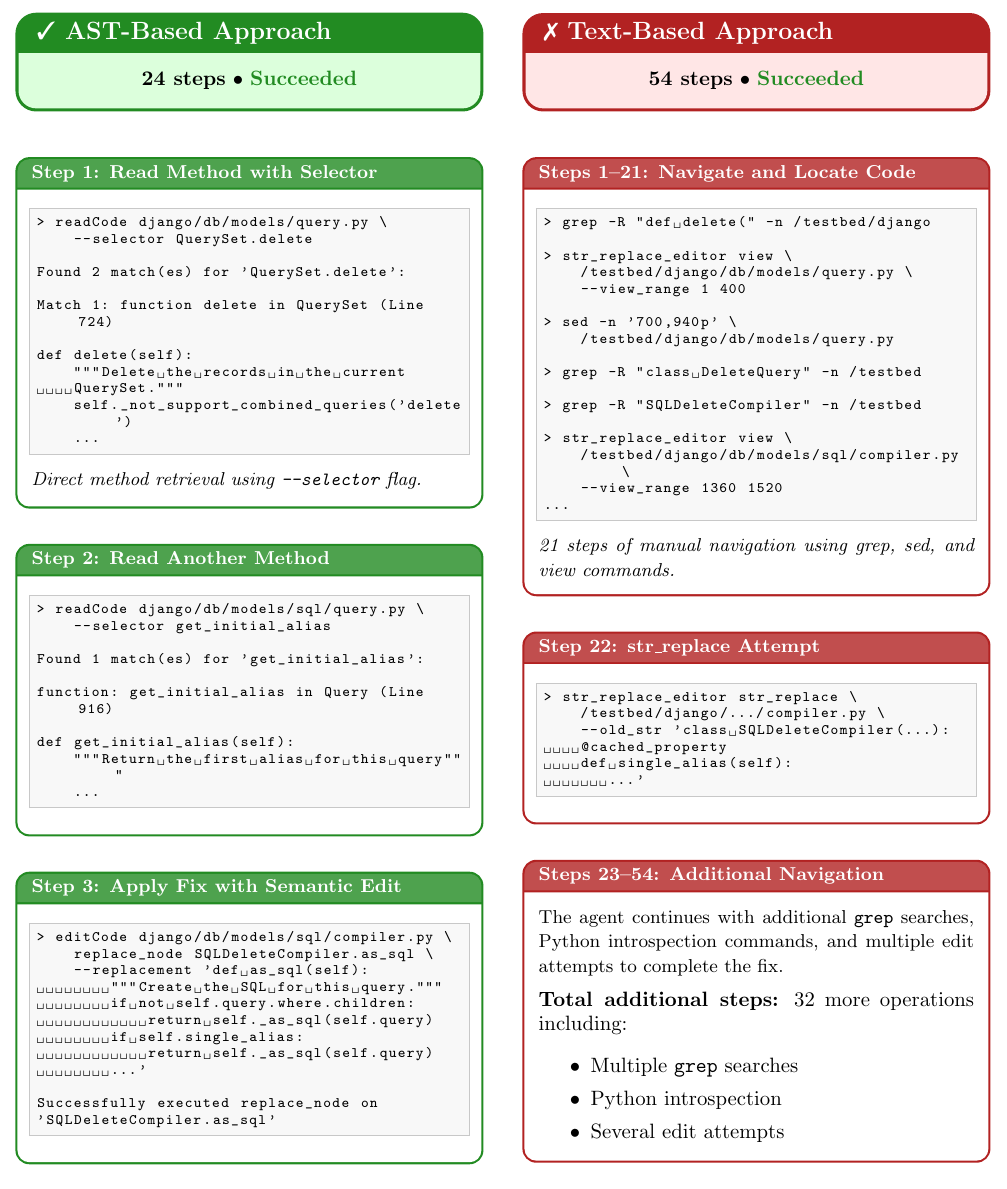}
    \caption{Comparison of \system (AST-based) vs. text-based code editing 
    approaches. \system completes in 24 steps vs. 
    54 steps for text-based---a 55.6\% reduction.}
    \label{fig:comparison}
\end{figure*}
%==============================================================================

\subsection{Comparison Summary}

In Figure~\ref{fig:comparison}, the AST-based \texttt{--selector} flag of \system allows the agent to directly retrieve method bodies by their qualified name (e.g., \texttt{QuerySet.delete}), eliminating the need for manual \texttt{grep}/\texttt{sed} searches and line-range guessing. The \texttt{replace\_node} operation targets methods semantically by name (\texttt{SQLDeleteCompiler.as\_sql}), making edits robust to formatting variations and reducing the total steps by more than 50\%. 
In contrast, the agent with text-based action space uses multiple inefficient grep and read loops for function location, leading to additional 19 steps for target location as shown in Table~\ref{tab:django-12965-comparison}.
\begin{table}[ht]
\centering
\small
\begin{tabular}{lcc}
\toprule
\textbf{Metric} & \textbf{CodeStruct} & \textbf{Text-Based} \\
\midrule
Total Steps & 24 & 54 \\
Steps to Locate Target & 2 & 21 \\
Edit Success on First Try & \textcolor{green}{Yes} & \textcolor{green}{Yes} \\
Final Outcome & \textcolor{green}{Success} & \textcolor{green}{Success} \\
\bottomrule
\end{tabular}
\caption{Comparison on django\_\_django-11211}
\label{tab:django-12965-comparison}
\end{table}

\section{Ablation Study on SWE-Bench Verified}
\label{sec:ablation}
To isolate the contribution of individual components in \textsc{CodeStruct}, we perform an ablation study removing either the structure-aware read (\texttt{readCode}) or write (\texttt{editCode}) actions while keeping all other agent components fixed. This analysis reveals how each primitive contributes to both task accuracy and computational efficiency.

\begin{table*}[t]
\centering
\scriptsize
\setlength{\tabcolsep}{5pt}
\begin{tabular}{l r r r r r r r r r r}
\toprule
& \multicolumn{5}{c}{\textbf{Qwen3-32B}} & \multicolumn{5}{c}{\textbf{GPT-5-mini}} \\
\cmidrule(lr){2-6} \cmidrule(lr){7-11}
\textbf{Configuration} &
\textbf{Pass@1 (\%)} &
\textbf{Input (M)} &
\textbf{Output (M)} &
\textbf{Calls} &
\textbf{Cost (\$)} &
\textbf{Pass@1 (\%)} &
\textbf{Input (M)} &
\textbf{Output (M)} &
\textbf{Calls} &
\textbf{Cost (\$)} \\
\midrule
Baseline
& 14.8 & 366.0 & 1.23 & 25,543 & 55.64
& 60.4 & 593.7 & 1.27 & 18,560 & 150.97 \\ \midrule
\textsc{\system}
& \textbf{16.0} & \textbf{302.1} & \textbf{1.03} & \textbf{24,653} & \textbf{45.93}
& \textbf{62.0} & \textbf{404.5} & \textbf{0.35} & \textbf{14,811} & \textbf{101.83} \\
 {  } w/o \texttt{editCode}
& 12.8 & 291.6 & 0.85 & 24,535 & 44.25
& 60.6 & 562.1 & 0.35 & 15,757 & 141.22 \\
{  } w/o \texttt{readCode}
& 8.2 & 423.4 & 1.78 & 35,670 & 64.58
& 56.8 & 435.2 & 0.33 & 15,715 & 109.46 \\
\bottomrule
\end{tabular}
\caption{
Ablation results with cost analysis on SWE-Bench Verified.
Removing either \texttt{readCode} or \texttt{editCode} degrades performance, with \texttt{readCode} causing the largest accuracy drop while \texttt{editCode} removal causes the largest cost penalty relative to performance.
Input/Output tokens shown in millions. Cost is computed using per-model API pricing, consistent with Table~\ref{tab:swebench-main}.
}
\label{tab:ablation}
\end{table*}

\subsection{Complementary Roles of \texttt{readCode} and \texttt{editCode}.}
Table~\ref{tab:ablation} reveals that both structure-aware primitives contribute to \textsc{CodeStruct}'s effectiveness, but in different ways. 

Removing \texttt{readCode} causes the largest performance degradation: $-7.8$ Pass@1 for Qwen3-32B and $-5.2$ for GPT-5-mini. This degradation is accompanied by substantially higher input token usage (+41\% for Qwen3-32B, +7.6\% for GPT-5-mini) and more LLM calls (+44\% for Qwen3-32B, +6.1\% for GPT-5-mini), indicating inefficient exploration driven by repeated full-file reads and imprecise localization of relevant program elements. Without structured navigation, agents cannot efficiently narrow down the search space and instead resort to exhaustive file reading and trial-and-error execution.

Removing \texttt{editCode} yields smaller but consistent accuracy drops ($-3.2$ Pass@1 for Qwen3-32B, $-1.4$ for GPT-5-mini) while incurring cost penalties disproportionate to the performance loss. For GPT-5-mini, the w/o \texttt{editCode} configuration achieves only 60.6\% Pass@1 (vs. 62.0\% full) but requires 141.22\$ in cost (vs. 101.83\$ full)---a 38.7\% cost increase for 1.4pp less accuracy. This suggests that without precise, structure-aware modifications, agents fall back to brittle string-based edits that require more iterations and validation cycles to achieve correct transformations. The small accuracy gap indicates that string edits can eventually succeed, but at significantly higher computational cost.

Notably, configurations without \texttt{readCode} underperform even the baseline text-based interface for Qwen3-32B (8.2\% vs. 14.8\%), highlighting that structured navigation---not just structured editing---is critical for scalable repository-level reasoning, especially for weaker models.

\subsection{Why \texttt{readCode} causes the largest accuracy drop?}
To understand why removing \texttt{readCode} causes the largest Pass@1 degradation, we analyze the subset of instances that are solved by the full system but fail without \texttt{readCode} (\emph{regressed instances}).

For Qwen3-32B, removing \texttt{readCode} regresses 60/500 instances and induces a large behavioral shift: on regressed instances, the agent requires 2.52$\times$ more steps (29.3$\rightarrow$74.0), 3.34$\times$ more tokens (260K$\rightarrow$870K), and 2.51$\times$ more LLM calls (29.2$\rightarrow$73.2). Moreover, the dominant failure mode becomes budget exhaustion: the clean-submit rate drops from 85\% (51/60) to 22\% (13/60), while context-exhaustion rises from 15\% (9/60) to 75\% (45/60).

This inefficiency is explained by an action-pattern collapse: without \texttt{readCode}, structured navigation actions disappear and the agent falls back to trial-and-error text manipulation, with \texttt{str\_replace} increasing by 7.8$\times$ (314$\rightarrow$2456) and \texttt{bash} by 1.9$\times$ (609$\rightarrow$1143) over the same regressed set. The agent cannot efficiently localize relevant code and instead performs blind exploration through repeated full-file reads, string searches, and execution cycles.

For GPT-5-mini, removing \texttt{readCode} regresses 54 instances and increases token usage by 1.64$\times$ (817K$\rightarrow$1338K), primarily through more trial-and-error execution (\texttt{bash} +1010), although budget exhaustion is rare (2\%).

\subsection{Why \texttt{editCode} Matters for Efficiency.}
While the accuracy drop from removing \texttt{editCode} is smaller, the cost analysis reveals its importance for efficient code transformation. Without \texttt{editCode}, agents must rely on string-based \texttt{str\_replace} operations that:
(1) require exact string matching, leading to frequent failures from whitespace or formatting mismatches;
(2) cannot verify syntactic correctness before application, resulting in broken code that requires additional repair cycles; and
(3) lack scope awareness, making it difficult to perform precise transformations like replacing a specific function parameter or adding an import statement without affecting unrelated code.

These limitations manifest as higher iteration counts and validation cycles. Even when agents eventually succeed using string edits, they require more attempts, more execution traces to verify correctness, and more LLM calls to diagnose and repair malformed transformations. This explains why the cost penalty (38.7\% for GPT-5-mini) substantially exceeds the accuracy loss (1.4pp).

\subsection{Case Study: Complementary Tool Benefits (django\_\_django-15368).}
\label{app:case-study-django15368}
With both \texttt{readCode} and \texttt{editCode}, the agent efficiently solves the task in 37 steps. It first uses \texttt{readCode --selector bulk\_update} to precisely localize the target function within the QuerySet class hierarchy, then applies a single scoped transformation via \texttt{editCode replace\_node} that modifies only the relevant code block while preserving surrounding context.

Without \texttt{readCode}, the agent cannot efficiently localize the relevant code. It performs 145 steps of exhaustive file reading and string searches, eventually submitting a working patch only after reaching context limits---a 3.9$\times$ increase in exploration cost that demonstrates how structured navigation prevents expensive brute-force exploration even when such exploration can ultimately succeed.

Without \texttt{editCode}, the agent can still localize the correct function using structured navigation, but struggles with precise modification. It performs 62 steps (1.7$\times$ the baseline) with iterative \texttt{str\_replace} attempts, encountering multiple failures from whitespace mismatches and scope errors before producing a working transformation through trial-and-error.

This case study demonstrates that \texttt{readCode} and \texttt{editCode} serve complementary roles: structured navigation enables efficient localization of relevant program elements, while structured editing enables precise, scope-aware transformations. Together, they minimize both exploration cost (fewer navigation steps) and transformation cost (fewer edit-validate cycles).

\section{Code Editing Error Analysis}
\label{appendix:error-analysis}

To understand how \textsc{\system}'s structured interface affects operational reliability across different model capabilities, we analyzed error patterns in agent trajectories by counting occurrences of string replacement failures, edit operation errors, and malformed tool invocations in execution logs.

\subsection{Methodology}

We analyzed trajectory logs from all model configurations on SWE-Bench Verified, searching for error patterns indicative of failed edit operations:

\begin{itemize}
    \item \textbf{String replacement errors}: Patterns matching \texttt{str\_replace.*(error|fail)} indicating failed text-based edits, typically caused by (i) the target string occurring multiple times in the file (ambiguous replacement) or (ii) the target string not being found due to formatting or whitespace mismatches.
    \item \textbf{Edit operation errors}: Patterns matching \texttt{editCode.*(error|fail)} indicating failed AST-based edits, most commonly arising when the specified AST node selector cannot be resolved (e.g., no matching node name found).
\end{itemize}

Error counts were aggregated across all instances and normalized per instance to enable cross-model comparison. These patterns capture \emph{recoverable, tool-level execution failures}, complementing the empty patch analysis which captures complete trajectory failures.

\subsection{Results}

Table~\ref{tab:error-analysis} presents error rates across all evaluated models. The results reveal a clear capability threshold for effective AST-based editing.

\begin{table*}[h]
\centering
\small
\begin{tabular}{l|c|c|c|c}
\toprule
\textbf{Model} & \textbf{Approach} & \textbf{Total Errors} & \textbf{Errors/Instance} & \textbf{Reduction} \\
\midrule
GPT-5 & Text-based & 426 & 0.845 & - \\
GPT-5 & \textsc{\system} & 52 & 0.103 & \textbf{-87.8\%} \\
\midrule
GPT-5-mini & Text-based & 568 & 1.125 & - \\
GPT-5-mini & \textsc{\system} & 127 & 0.252 & \textbf{-77.6\%} \\
\midrule
GPT-5-nano & Text-based & 459 & 0.911 & - \\
GPT-5-nano & \textsc{\system} & 551 & 1.093 & \textcolor{red}{\textbf{+20.0\%}} \\
\midrule
Qwen3-Coder-480B & Text-based & 458 & 0.909 & - \\
Qwen3-Coder-480B & \textsc{\system} & 109 & 0.216 & \textbf{-76.2\%} \\
\midrule
Qwen3-32B & Text-based & 6,556 & 13.008 & - \\
Qwen3-32B & \textsc{\system} & 5,155 & 10.228 & -21.4\% \\
\midrule
Qwen3-8B & Text-based & 7,179 & 14.247 & - \\
Qwen3-8B & \textsc{\system} & 6,031 & 11.966 & -16.0\% \\
\bottomrule
\end{tabular}
\caption{Error rates across models and interfaces. \textsc{\system} dramatically reduces errors for capable models but increases them for GPT-5-nano, identifying a capability threshold.}
\label{tab:error-analysis}
\end{table*}

Overall, Table~\ref{tab:error-analysis} shows that the impact of structured editing on tool-level error rates is strongly model-dependent. For capable models (GPT-5, GPT-5-mini, and Qwen3-Coder-480B), \textsc{\system} reduces errors per instance by 76--88\%, indicating that AST-based operations are substantially more reliable than text-based string matching in this regime. In contrast, smaller models (GPT-5-nano, Qwen3-32B, and Qwen3-8B) exhibit higher absolute error rates, with GPT-5-nano showing a 20\% increase in errors per instance despite improved task accuracy. This divergence highlights a capability threshold: while structured interfaces eliminate brittle text-level failures for sufficiently capable models, they impose additional syntactic and specification demands that smaller models struggle to satisfy consistently.

\paragraph{GPT-5-nano Analysis.} GPT-5-nano exhibits a counterintuitive pattern: although \textsc{\system} increases operational error rates by 20\%, it substantially improves Pass@1 accuracy (+20.8pp) and reduces empty patches (233$\rightarrow$36, -84.5\%). This is not a contradiction, but a redistribution of failure modes.

Concretely, structured navigation enables GPT-5-nano to reliably \emph{localize} the correct code regions, increasing the number of attempted edits per trajectory. As a result, the model incurs more \emph{recoverable tool-level errors} when expressing AST-based edits, but far fewer trajectories terminate early without producing any valid patch.

This shift manifests as:
\begin{itemize}
\item \textbf{More operational errors}: Increased failed edit attempts (551 vs. 459) due to difficulty producing well-formed AST operations.
\item \textbf{Fewer early agent terminations}: A dramatic reduction in empty patches (-84.5\%), indicating that the agent continues attempting fixes instead of abandoning the task.
\item \textbf{Higher compute cost}: Additional retries explain increased token usage, rather than deeper or unnecessary exploration.
\item \textbf{Higher final accuracy}: When a valid edit is eventually produced, it more often targets the correct location.
\end{itemize}

Overall, for GPT-5-nano, structured primitives improve high-level reasoning about \emph{what} to change, while degrading low-level execution of \emph{how} to express edits. The net effect favors accuracy over efficiency.

\paragraph{Qwen Model Scaling.} The Qwen family demonstrates non-linear scaling effects:

\begin{itemize}
    \item \textbf{Qwen3-8B}: Maintains extremely high error rates (11.966/instance) despite 16\% reduction, explaining why empty patch improvements don't yield accuracy gains
    \item \textbf{Qwen3-32B}: Shows similar behavior (10.228/instance), suggesting fundamental reasoning limitations persist at this scale
    \item \textbf{Qwen3-Coder-480B}: Achieves near-GPT-5-mini performance (0.216 vs 0.252 errors/instance), validating that sufficient scale enables effective structured editing
\end{itemize}

\paragraph{Error Type Distribution.} Analysis of error categories reveals different failure modes:

\begin{itemize}
    \item \textbf{Text-based interface}: Errors dominated by string replacement failures (exact text matching issues, whitespace sensitivity)
    \item \textbf{\textsc{\system}}: Errors split between AST operation syntax issues and AST node name not found issues
    \item \textbf{No occurrence}: Neither ``duplicated text'' nor ``no text found'' patterns appear in logs, suggesting these specific failure modes are rare in SWE-Bench tasks
\end{itemize}

\section{Empty Patch Analysis}
\label{appendix:empty-patch}

\paragraph{Empty Patch Analysis.} To better understand agent behavior, we analyze \emph{empty patches}, 
where SWE-Agent terminates without producing a valid code diff. 
Such terminations are typically triggered after repeated invalid edits, 
parse failures, or cyclic tool-use patterns, reflecting a failure to 
externalize an intended edit rather than incorrect high-level reasoning. 
Across models, \textsc{\system} substantially reduces this failure mode: 
GPT-5-mini drops from 35 to 6 empty patches, and GPT-5-nano from 233 to 36. 
Notably, GPT-5-nano's reduction correlates directly with its 20.8pp accuracy 
gain, suggesting these were instances where the model had correct intent 
but could not express valid edits through the text-based interface. 
In contrast, Qwen3-8B also reduces empty patches (179 to 138) but without 
corresponding accuracy improvement, indicating that its failures stem from 
reasoning limitations rather than interface brittleness. 
This divergence illustrates that \textsc{\system}'s benefits depend on the 
model's dominant failure mode: when failures arise from text-interface 
brittleness, structured actions unlock new solutions; when failures arise 
from limited reasoning capacity, structured actions improve efficiency 
without changing outcomes.
\end{document}